# ChatSOS：基于大语言模型的安全工程知识问答系统

唐海洋， 刘振翼 副教授， 陈东平 副教授， 初庆钊 副教授

（北京理工大学 机电学院，北京 100081）

**【摘 要】**近年来，大语言模型在自然语言对话任务中取得了显著进展，同时在安全领域的应用也呈现出卓越的潜力。然而，这些模型在特定领域的任务处理中受到语料库规模、输入处理能力、隐私性等多方面限制。为此，提出了一个基于大语言模型的安全工程知识问答系统，旨在提高模型的问题理解和解答能力。采用提示学习的方法，整合外部知识库作为大语言模型的知识补充，给大模型提供可靠、可更新的信息源。知识系统以往年事故调查报告为基础，进行统计分析，并将相关信息编码为向量，构建一个向量数据库，实现对问题的相关性检索。研究表明，这种整合外部知识的大语言模型具备深入分析问题、自主任务分配的能力，能够对事故报告进行详尽总结和建议。该研究提出的外部知识整合方法极大地提高了大语言模型解决安全工程专业问题的能力，为大语言模型在安全工程领域的可靠应用奠定了基础。

**【关键词】**安全工程；大语言模型；模型代理；向量数据库；提示工程

# ChatSOS: LLM-based knowledge Q&A system for safety engineering

TANG Haiyang, LIU Zhenyi, CHEN Dongping, CHU Qingzhao

(School of Mechatronic Engineering, Beijing Institute of Technology, Beijing 100081, China)

**Abstract:** Recent advancements in large language models (LLMs) have notably propelled natural language processing (NLP) capabilities, demonstrating significant potential in safety engineering applications. Despite these advancements, LLMs face constraints in processing specialized tasks, attributed to factors such as corpus size, input processing limitations, and privacy concerns. Obtaining useful information from reliable sources in a limited time is crucial for LLM. Addressing this, our study introduces an LLM-based Q&A system for safety engineering, enhancing the comprehension and response accuracy of the model. We employed prompt engineering to incorporate external knowledge databases, thus enriching the LLM with up-to-date and reliable information. The system analyzes historical incident reports through statistical methods, utilizes vector embedding to construct a vector database, and offers an efficient similarity-based search functionality. Our findings indicate that the integration of external knowledge significantly augments the capabilities of LLM for in-depth problem analysis and autonomous task assignment. It effectively summarizes accident reports and provides pertinent recommendations. This integration approach not only expands LLM applications in safety engineering but also sets a precedent for future developments towards automation and intelligent systems.

**Keywords:** safety engineering; large language model; Agent; vector database; prompt engineering

## 0 引 言

在如今信息技术快速发展的背景下，知识的获取和管理受到前所未有的巨大冲击与挑战。信息获取的渠道增多，人们已经不再局限于传统的图书或基础互联网检索，人工智能技术的发展带来了便捷的方式，但同时信息的真实性和完整性都更难以得到保证。对于事故调查报告，受到保密性和信息复杂程度的限制，目前的相关信息的收集和管理还极度依靠人工，传统的关系型数据库管理系统（Relation Database Management System, RDBMS）的维护也需要依赖专业人员[1]，导致这种传统的信息检索方式无法满足迅速准确获取知识的需求，成为安全事故调查领域突出的痛点。

在这一背景下，GPT[2-4]，LLaMA[5-7]，PaLM[8]等大语言模型（Large Language Model, LLM）在自然语言处理（Natural Language Processing, NLP）任务中的快速发展引人注目，在理解和响应人类指令方面能力突出。然而，LLM 在特定领域的应用，如事故调查报告的收集和管理，还面临着诸如信息滞后、语料缺乏和模型自身"幻觉"的挑战，其在垂直行业内的应用并不理想。为了应对这些问题，部分研究者已经开始探索通过模型微调（Fine Tuning, FT）和知识注入来改善 LLM 在专业领域的表现，CUI J 等[9]采用 LoRA 方法[10]微调关键词提取模型，从用户查询中提取法律相关关键词。此外，他们还使用了丰富的法律案例文本微调 LLaMA 模型，以此创建了为法律咨询场景设计的 LLM(ChatLaw) VAGHEFI S A 等[11]通过为模型提供对 IPCC 第六次评估报告（AR6）的访问权限，显著提升了模型在回答准确性方面的性能，展示了 LLM 作为未来化学助手的潜力；BRAN A M 等[12]为 LLM 集成了 13 种专家设计的工具，提高了 LLM 在化学领域的性能。

基于对安全领域数据库的持久健康发展的考虑，本工作研究并建立了相关知识库；针对基座大模型在专业领域内问答能力不足的问题，提出一个结合大语言模型和向量数据库的安全事故数据管理分析方法，使用向量检索的方法提高了检索效率和准确性，使用提示模板针对性地改善了大模型易出现的幻觉（hallucination）问题，实现了一个基于大语言模型的安全工程知识问答系统（ChatSOS）。该智能问答系统支持对安全事故相关提问的快速响应，对事故报告进行深入的总结与分析，并能提取关键数据用于进一步处理。ChatSOS 为安全事故报告的自动化处理提供了新的可能性，促进了安全管理体系技术创新。

## 1 基本理论

### 1.1 大语言模型

大语言模型是目前人工智能领域内的研究热潮，是属于自然语言处理领域内的一个分支，LLM 通过在大量文本上训练，以捕捉语言的统计学规律并自动调整模型参数。语言模型本质上是进行一系列的概率计算，对给定文本序列的合理性进行分析并量化，其中 N 元语法模型是概率语言模型的基础，预测过程按下式计算[13]：

$$p(w_1, \dots, w_2) = \prod_{i=1}^{n} p(w_i|w_1, \dots, w_{i-1}) \quad (1)$$

$$p(w_i|s) = \frac{C(s,w_i)}{C(s)} \quad (2)$$

其中，$p(w_1, \dots, w_2)$表示词序列出现的概率，$w_i$表示词序列$[w_1, \dots, w_N]$中的一个单词，$p(w_i|s)$表示在给定上下文$s$的条件下中某一单词$w_i$出现的概率估计，$C(s, w_i)$表示为上下文$s$与单词$w_i$共同出现的次数。模型通过特定训练和调优过程后，本身可以具备一定的思维链（Chain Of Thought, CoT）能力，能将复杂任务拆分为多个可解决的中间步骤。在 ChatSOS 中，相应的大语言模型主要作为生成式能力的核心，使用提示词激发模型的推理能力，引导生成结果的大致方向。

### 1.2 数据库系统

传统的 RDBMS 如 SQL 数据库，以表格的形式存储数据，可保证数据的一致性和完整性，但对于数据的形式要求比较严苛（结构化数据），难以提取非结构化数据（如自然语言）中的信息进行储存和分析，这会增加对有效信息提取与保存的难度。同时 RDBMS 在处理大规模，高纬度数据的表现不够理想。向量数据库通常会使用到"文本-向量"编码模型，将文本块编码到高维度的向量空间中，在向量嵌入过程中，语义相似的文本片段在向量空间中位置相近，保证了向量数据库检索时的泛化能力。

对于知识问答系统而言，语言模型作为内核，数据多为报告与总结形式的非结构化自然语言文本，且数据量通常比较大。因此，采用向量数据库能有效规避 RDBMS 对数据形式的严格要求和其在处理大规模、高维度数据性能不足的缺陷，提高数据处理和分析的效率。

### 1.3 提示工程（Prompt Engineering）

对于不同研究团队推出的基座大模型，根据模型训练语料和对齐标准的不同，模型的各项能力指标（如泛化能力）也会有区别。如 ChatGPT-3.5 的训练语料库截止于 2021 年 9 月，其对于此时间点后发生的事件无法作出回答，或是出现大模型幻觉。

这本质上是源于训练数据的不足，除了数据时效性的问题，专业领域数据的敏感性也是导致这一问题的另一大原因。

为解决 LLM 训练语料不足的问题，最大程度保证回答的真实性与准确性，采用提示学习的方法，注入相关知识片段，并将预测目标与具有对应语义关系的词句相联系，整合到用户提问中辅助回答。这种方式并不会对模型本身的参数造成影响，而是间接利用模型的逻辑推理能力和一定的泛化能力。相比直对模型微调，提示学习的成本相对较低，效果显著且结果基本可控。本文所提到的提示模板主要采用将预设场景和相关知识注入问答模板的方式进行设计。

## 2 ChatSOS（Chat Security Oracles）

本工作以 LLM 模型作为生成式内容核心，基于提示学习的方法，使用向量数据库与 MySQL 数据库相结合，分别以“向量+标识符”和“标识符+文本”的形式储存数据，构建了一套包含知识存储、提取与注入的系统 ChatSOS。

### 2.1 系统框架

通过分析系统不同部分的功能需求，基于提示学习和数据库建设技术构建知识问答系统架构，如图 1 所示。该系统包括 3 个基本层级，包括输入层，运算层和输出层。

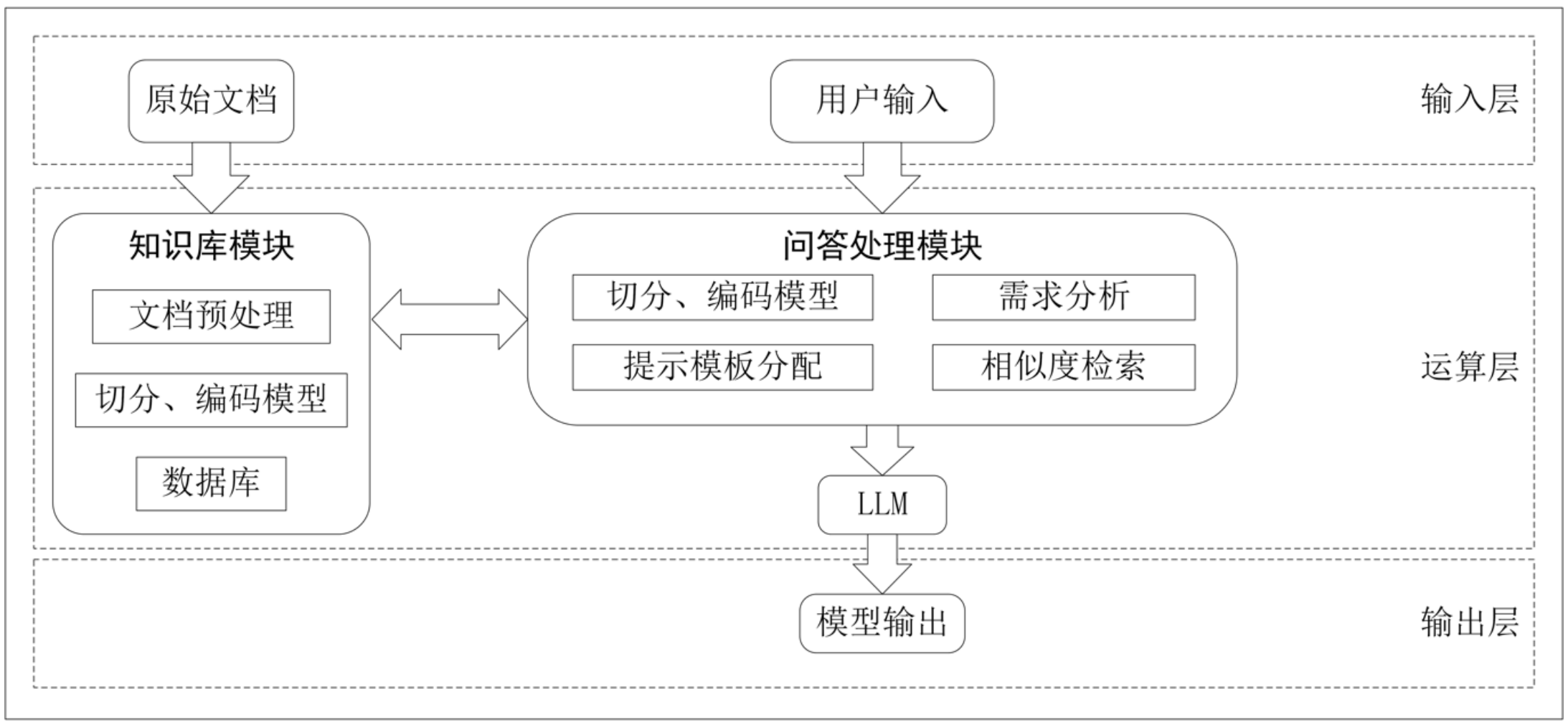

**图 1 基于提示学习和数据库操作的知识问答系统架构**

**Fig.1 Framework of the knowledge Q&A system based on prompt learning and database operation**

运算层部分可细分为两大模块：知识库和问答处理。在知识库模块的构建中，系统对原始文档执行格式标准化处理，确保文档格式的一致性；通过文本预处理和分割，将文档拆分成可管理的片段；“文本到向量”的转换过程中，通过使用预训练的嵌入模型将文本数据映射到高维向量空间，以便捕捉语义信息。每个生成的向量及其对应的文本片段都会被分配一个唯一标识符。向量存储在专门的向量索引系统 Milvus 中，以支持高效的相似性检索，而文本片段及其元数据则存储在关系型数据库 MySQL 中，以便进行结构化查询与数据处理。在问答处理过程中，构建代理（Agent）来处理信息检索、需求分析、问题拆解、模型调用、用户交互等方面的问题。大语言模型容易产生误导性和“幻觉”（Hallucination）现象，主要是因为缺乏真实可靠的语料或使用了误导性的训练语料。因此，我们通过分析不同用户群体的需求（如表 1 所示），针对性地构建了提示模板，这些提示模板通过知识注入和定向引导，有助于优化模型的输出。初步实验表明，这种方法能有效减少模型在处理不熟悉问题时生成不准确或与偏离事实信息的频率。

**表 1 行业需求分析**

**Tab.1 Industry demand analysis**

| 场景 | 需求 | 具体形式 |
|---|---|---|
| 政府部门 | 总结、报告、通知 | 政府公文 |
| 科研工作者 | 数据分析、研究成果分享 | 学术论文、研究报告、学术演讲 |
| （本专业）企业从业者 | 安全风险评估、事故预案 | 安全风险评估报告、安规、安全培训资料 |
| （跨专业）企业从业者 | 跨部门协作、了解紧急应对措施 | 跨部门协作指南、应急相应指南 |
| 公众 | 公共安全教育、灾害应对 | 科普文章、新闻稿件 |

### 2.2 知识问答处理流程

图 2 展示了 ChatSOS 在处理知识问答任务时的工作流程。在数据处理过程中，用户查询首先通过 BAAI general embedding (BGE)模型[14]编码为向量表示，此后这一向量表示被用于向量数据库中的相似度检索。根据预设的算法，数据库返回相似度得分最高的前 *k* 个结果。得益于向量保留的丰富语义信息和高效的向量检索算法，ChatSOS 在处理多样化和复杂的用户查询时表现出卓越的泛化能力，图 3 展示了利用 t-SNE 技术[15]从高维向量空间降维至二维平面的结果，通过数据点之间的相对位置和集群趋势反映出数据点在语义上的关联性。相同的颜色代表其有类似的语义。与此同时，模型代理接管问答任务并解析用户查询，根据不同场景需求选择合适的提示模板。在接收到数据库的查询结果后，模型代理会将这些结果作为补充知识融入提示模板中。最终，这些整合后的提示词被输入到大型模型中，系统随后将模型的输出结果反馈给用户。

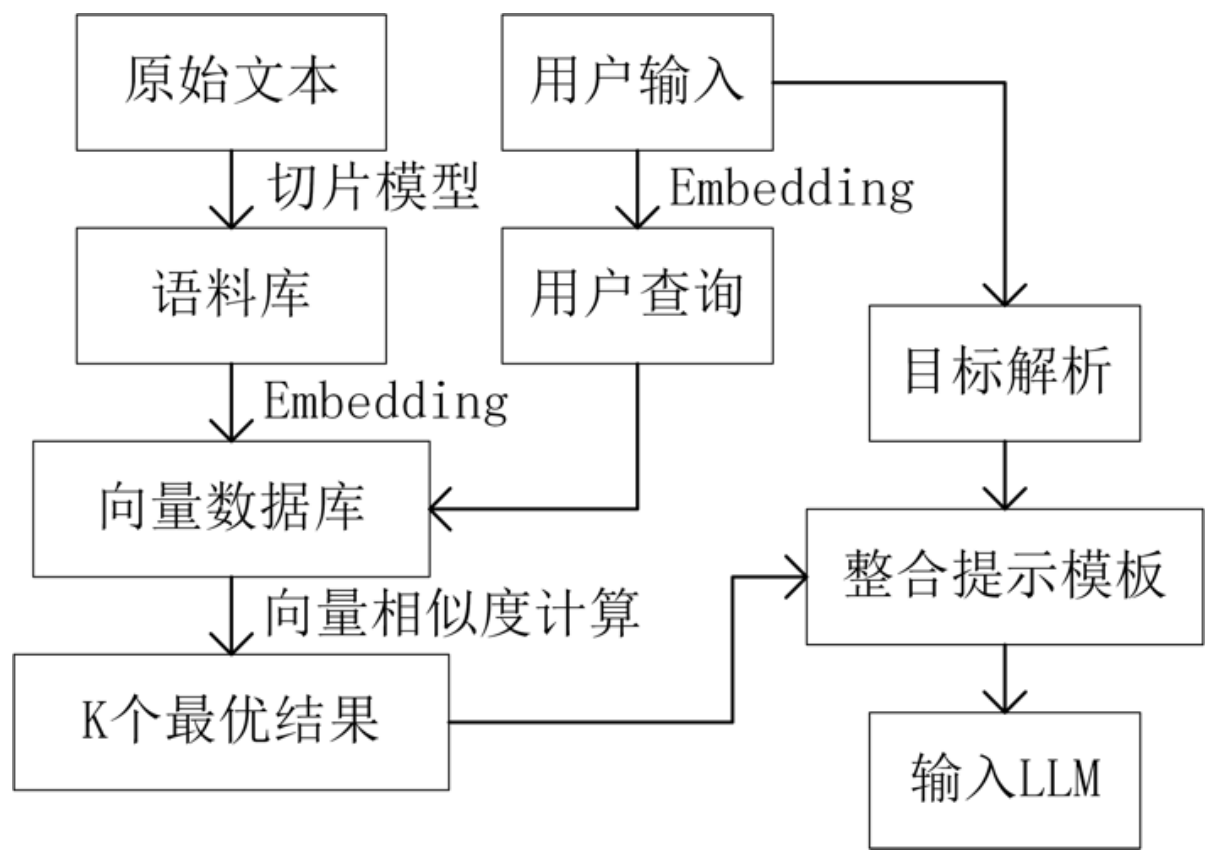

**图 2 ChatSOS 问答任务处理流程**

**Fig.2 ChatSOS Q&A processing workflow**

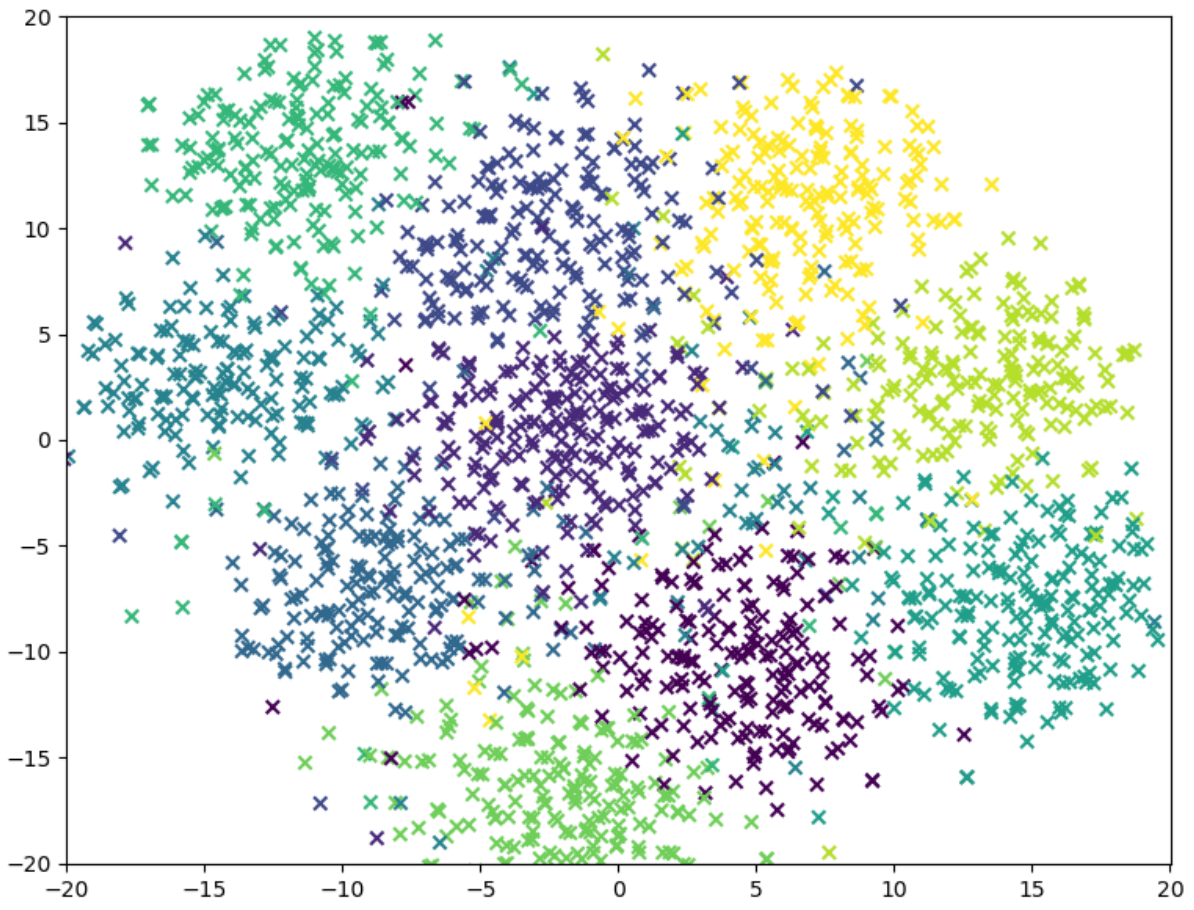

**图 3 数据点的集群分布**

**Fig.3 Cluster distribution of data points**

## 3 对比测试

### 3.1 评价标准

针对现有对大语言模型的常用评价标准研究，结合实际用户使用需求，我们为基于知识库的问答系统设计了一套评分标准，如表 2 所示。

**表 2 系统问答响应评分表**

**Tab.2 Rating chart for system Q&A response**

| 评分标准 | 权重 | 分数（0-5） |
|---|---|---|
| 准确性 | 0.3 | |
| 可靠性 | 0.3 | |
| 适应性 | 0.2 | |
| 简洁性 | 0.1 | |
| 快速性 | 0.1 | |
| 加权总分 | 1 | |

该评分表从五个维度对 ChatSOS 输出进行评分：准确性评判回答问题的准确程度；可靠性评判系统回答的一致性；适应性评判系统对问题要求作出的调整；简洁性评判回答是否冗长且直接相关；快速性则评价系统的响应速度。

### 3.2 提示模板构建

本研究分析不同人群需求，通过针对性地构建提示模板，采用"*n*-shot learning"的方式，实现了适应性强的多样化提示方法，其中 *n* 表示问答模板数。表 3 举例说明针对不同场景需求设计提示模板的基本流程。

表 3 提示模板设计

Tab.3 Prompt template designing

| | |
|---|---|
| 场景预设<br>&<br>知识注入 | 你是专业的 AI 助手，根据给定信息回答用户提问。回答时逐步思考，若依给定知识无法回答，只需说明给定信息不足，绝对禁止编造。已知信息(information)，用户提问(query)。 |
| 事故分析 | 执行分析事故报告任务，从“事故概况、事故原因、责任分析、事故影响及预防、总结与建议”等多方面合理分析。 |
| 问答模板 | Q1:请分析 2007 年湘西凤凰塌桥事故 |
| | A1:<br>**1.事故概括**：时间-2007 年 8 月 13 日；地点-湖南省湘西凤凰县；环境条件-堤溪沱江大桥建设中；事故等级-严重责任事故；伤亡-64 人死亡，22 人受伤；直接经济损失-3974.7 万元。<br>**2.责任分析**：调查认定施工、建设单位严重违规，监理单位、质量监督部门失职，湘西自治州和凤凰县两级政府及湖南省交通厅、公路局监管不力。<br>**3.事故原因**：大桥主拱圈砌筑材料未满足规范和设计要求，施工工序不合理，主拱圈砌筑质量差，降低了结构整体性和强度，随着施工荷载增加导致坍塌。<br>**4.事故影响**：人员伤亡及直接经济损失。<br>**5.事故预防**：提升工程管理和现场监管，完善设计审查流程，强化责任意识等。 |

### 3.3 效果对比

依照 3.1 节提出的标准，我们设计了一系列针对 ChatSOS 系统的测试项目，包括通识性问答、格式规范任务和具体事故分析。测试中，将 ChatSOS 与 GPT 3.5-Turbo 模型的输出结果进行了对比分析（表 4），其中 ChatSOS 通过 API 接口调用 GPT3.5-Turbo 作为其内核的大型语言模型，以验证我们所提出的改进结构对问答任务提升的有效性。

依照评分标准对 ChatSOS 与 ChatGPT 进行打分，各项评分对比如图 4 所示。在处理通识性问题的问答任务中，由于训练语料充足，ChatGPT 的回答覆盖面较广，确保了较高的准确度和可靠性。在讨论燃气泄漏的具体原因及预防措施时，ChatGPT 进行了详尽的分类，例如将泄漏原因细分为设备故障和人为疏忽等多个方面。另一方面，ChatSOS 凭借其知识库的深入整合，能够在提供回答时援引具体数据和模板，提供了实际案例，例如在回答中提到由于动物啃咬或车辆碰撞引起的燃气管道损伤，从而在广泛性与细致性之间实现了有效的平衡。

在处理符合特定格式规范的任务中，ChatGPT 表现出较好的适应性，能准确理解政府通知的格式要求，将内容细分成多个条目进行总结。然而，由于缺少具体数据，ChatGPT 的回答出现了所谓的“模型幻觉”，在总结“太阳岛洗浴中心事故”的基本情况时给出错误的事故时间，在分析事故原因时未能明确指出具体的致因因素，而是给出了一些较为通用的模糊表述，且在之后处理措施部分的总结中也出现了类似的模糊回答。另一方面，ChatSOS 在输出结果中准确地描述了事故的基础情况，并在事故原因分析中细化到了桑拿炉的安全问题，其余关键点如责任追究、隐患排查、救援及后续处理工作等也展现了较为全面的考量。此外，在未提供具体负责机构名称的情况下，ChatSOS 凭借其深度的信息理解和推理能力，通过现有数据推导出了具体的负责机构为驻马店市政府，也再次强化了其输出结果的可信度。

在具体事故分析任务中，ChatGPT 的表现受限于其训练语料的局限性，未能提供详细的事故分析。ChatSOS 在分析“山东四方安装工程公司较大爆燃事故”时，通过提示工程注入相关知识条目和问答模板，实现对事故的全方位分析，ChatSOS 不仅提供了事故的全面概述，还对事故涉及的各方，包括四方公司、施工单位、东营港经济开发区管委会及相关监管部门都进行了详尽的责任分析。此外，它还能够区分事故的直接原因与间接原因，评估事故的影响并提出预防策略。ChatSOS 在处理分析任务时具备深度和广度，充分证明了其在复杂任务处理上的优势。

**表 4　ChatSOS 综合测试**

**Tab.4　Performance evaluation of ChatSOS**

| 用户查询 | ChatSOS | GPT 3.5-Turbo |
|---|---|---|
| **（通识性问答）**<br>为什么会发生燃气泄漏，有什么预防办法吗? | 根据给定的信息，我们可以得知以下情况可能导致燃气泄漏：<br>1.燃气管道老化腐蚀：老旧燃气管道设施腐蚀造成的事故比例增加，这类事故隐蔽性强、不易及时发现，易酿成严重后果。<br>2.管道破损：管道的老化破损、脱落或动物咬噬等情况可能导致燃气泄漏。<br>3.第三方施工破坏：燃气管网中，第三方施工破坏事故数量占比较高，可能施工不慎损坏了燃气管道。<br>4.车辆撞击：地上燃气管道防护措施不到位，车辆撞击导致燃气管道泄漏。<br>针对燃气泄漏的预防措施包括：<br>1.管道维护与更新：加强老旧燃气管道的更新改造，及时检修和更换老化腐蚀的管道，减少泄漏风险。<br>2.定期检查：定期对燃气管道进行检查和维护，确保管道的完整性和安全性。<br>3.连接软管和燃气灶具的检查：用户端连接软管问题仍然是引发户内事故的主要原因，应定期检查连接软管和燃气灶具的安全性，及时更换老化损坏的软管和灶具。<br>4.燃气阀门的正确使用：用户在使用燃气时，应正确关闭燃气阀门，避免瓶阀未关闭引发事故。<br>5.燃气灶具的正确使用：用户在使用燃气灶具时，应避免无人看管和干烧，加强对用户安全使用燃气知识的宣传教育，推广使用具有防干烧功能的灶具。<br>6.加强安全宣传教育：对用户和相关从业人员进行燃气安全宣传教育，提高他们的安全意识和使用燃气的技能。<br>综上所述，通过管道维护与更新、定期检查、正确使用燃气阀门和灶具、加强安全宣传教育等措施，可以预防燃气泄漏事故的发生。 | 燃气泄漏可能由多种原因引起，包括设备故障、人为错误、管道损坏、不当使用或安装等。以下是 一些常见的燃气泄漏原因和预防措施:<br>**燃气泄漏原因:**<br>1.设备故障或老化:燃气设备、管道或阀门因长时间使用或制造缺陷可能出现故障，导致泄漏。<br>2.人为错误:错误的操作、不当维修、不规范的安装和施工可能导致管道破裂或连接处松动， 引发泄漏。<br>3.管道损坏:管道遭受外部压力、物理损坏、地震或自然灾害可能导致管道破裂。<br>4.不当使用:例如，在不通风的空间使用燃气器具可能导致积聚的燃气泄漏。<br>**燃气泄漏的预防和安全措施:**<br>1.定期检查和维护:定期检查家庭用具、管道和燃气设备，确保其状态良好，没有泄漏。<br>2.合格安装:由专业人员合格安装和维护燃气设备，确保安装符合相关安全标准。<br>3.使用燃气报警器:在可能泄漏的区域安装燃气报警器，能够及时检测到燃气泄漏并发出警报。<br>4.通风:确保室内通风良好，特别是在使用燃气设备的封闭空间内。<br>5.安全使用:使用燃气器具时遵循使用说明，避免长时间不间断使用，确保炉具在使用后关闭。<br>6.防止损坏:避免在地下或易受损害的地方铺设燃气管道，以减少外部压力引起的破裂风险。<br>7.应急处理:如果怀疑有燃气泄漏，立即打开门窗通风，离开泄漏现场，不要使用手机或电器 设备，以免引发火源，然后拨打紧急救援电话。<br>请注意，如果你怀疑家中有燃气泄漏，绝对不能使用明火、电器开关，或者手机等可能引发火源的 设备，应立即采取安全措施并呼叫紧急救援电话。 |
| **（格式规范）**<br>按照政府公文的写作规范，发布一份关于太阳岛洗浴中心事故的政府事故调查结果通知。 | 【政府事故调查结果通知】<br>尊敬的各相关方：<br>根据《生产安全事故报告和调查处理条例》，经过对驻马店市开发区太阳岛洗浴休闲会馆事故进行全面调查，现将调查结果通知如下：<br>**一、事故概况：**2021年11月18日16时57分，太阳岛洗浴休闲会馆男浴区桑那房内的桑那炉发生爆炸，造成3人死亡，3人受伤，直接经济损失约500万元。<br>**二、事故调查经过：**1.事故发生前，太阳岛会馆正处于翻新装修施工的尾声，计划于11月19日开业。2.事发当日，工作人员某某[1]调试设施设备时，打开了男浴区桑那炉控制开关，并调试温度在65度-68度之间自动控温。3.然而，由于未及时发现和消除桑拿炉存在的安全隐患，导致爆炸事故的发生。<br>**三、事故原因分析：**1.太阳岛洗浴休闲会馆在装修施工期间未能严格按照安全生产规定进行管理，未落实安全生产主体责任，未督促指导沐浴场所落实安全管理规定。2.太阳岛洗浴休闲会馆未能加强沐浴行业安全日常检查，未及时消除事故隐患。3.太阳岛洗浴休闲会馆对桑拿炉存在的安全隐患未能及时发现和处理，导致爆炸事故的发生。<br>**四、责任追究：**根据调查结果，对驻马店市开发区太阳岛洗浴休闲会馆的管理部门、相关责任人员存在管理不到位、监管不力等问题，将依法进行相应责任追究。<br>**五、安全隐患整改：**驻马店市开发区将加强对沐浴行业的管理工作，督促沐浴场所落实安全生产主体责任和相关安全管理规定，加强日常检查，及时消除事故隐患，确保公众的安全和利益。<br>**六、救援和善后工作：**事故发生后，市委市政府成立了事故抢险救援指挥部，组织了现场救援、调查和善后处置工作，全力排查安全隐患，防范发生次生事故，同时组织医疗救治工作，确保受伤人员得到及时救治。<br>**七、倡议：**广大市民和相关单位加强安全意识，严格按照安全生产规定进行管理，共同维护社会的安全稳定。<br>特此通知。<br>驻马店市政府 日期：XXXX年XX月XX日 | 太阳岛洗浴中心事故的政府事故调查结果通知<br>[日期]<br>尊敬的市民：<br>根据国家法律法规和《政府信息公开条例》，为向社会公众及时、公正、公开地通报太阳岛洗浴中心事故的调查结果，将事故详细调查情况通告如下：<br>**一、事故基本情况**<br>2022 年月日，太阳岛洗浴中心发生火灾事故。事故发生后，市政府立即启动应急响应，组织消防、安监、公安等相关部门全力展开救援和调查工作。<br>**二、事故原因**<br>经过调查，事故的直接原因是洗浴中心内部设施设备存在安全隐患，加之管理不善，未能采取有效措施加以防范。具体原因包括（但不限于）：设备老化、维护不善、员工培训不足、应急预案不健全等。<br>**三、事故处理措施**<br>1. 责任追究：对相关责任人进行严肃问责，追究其领导责任和监管责任。<br>2. 安全整改：对全市范围内的洗浴、娱乐等场所进行全面排查，督促整改存在的安全隐患。<br>3. 加强监管：加大对公共场所的日常监管力度，确保市民生活、娱乐场所的安全。<br>4. 加强宣传教育：加强火灾、安全知识的宣传，提高市民的自防能力。<br>**五、结语**<br>政府将一如既往地高度重视人民群众的生命财产安全，对于类似事故将继续加大监管力度，切实保障市民的生命和财产安全。感谢广大市民对政府工作的理解和支持，也请大家加强自身安全意识，共同维护社会安定与和谐。<br>特此通知。<br>[政府机关名称] [日期] |
| （事故分析）请简要分析一下山东四方安装工程公司较大爆燃事故 | **事故概括：**山东四方安装工程公司在东营港经济开发区瑞林公司的碳钢系统污油罐改造施工过程中发生较大爆燃事故。事故发生于 2022 年 7 月 26 日上午 9 时 30 分，造成 3 人死亡，2 人轻伤，直接经济损失约 512.8 万元。<br>**责任分析：**事故中，山东四方安装工程公司未落实施工单位安全生产主体责任，对施工队管理不到位。东营港经济开发区管委会及其经济发展局、应急管理局也未认真履行安全生产监管职责。<br>**事故原因：直接原因**是四方公司施工人员在污油罐存有污油、罐顶中心透气孔敞开的情况下，超出《特殊动火安全作业证》许可作业范围，违规在罐顶中心透气口附近进行焊接作业。**间接原因**包括四方公司安全管理混乱、安全生产责任制严重不落实、隐患排查治理不到位以及对外来施工队伍管理不到位等。<br>**事故影响：**事故造成 3 人死亡，2 人轻伤，直接经济损失约 512.8 万元。此外，事故还对周边环境造成了一定的影响。<br>**事故预防：**为预防类似事故的发生，可采取以下措施：加强施工单位的安全生产责任制落实，完善安全管理制度和操作规程；加强隐患排查治理工作，确保施工现场的安全条件；加强对外来施工队伍的管理，确保其安全生产条件符合要求；加强安全风险分析，确保动火作业的风险辨识和管控措施到位。<br>**爆炸威力分析：**根据事故描述，污油罐发生了爆燃，罐体撕裂。具体的爆炸威力需要进一步的详细分析和调查。 | 很抱歉，根据我截止到 2021 年 9 月的知识，我无法提供有关"山东四方安装工程公司较大爆燃事故"的详细信息。请注意，由于我无法获取实时信息，也无法提供截至目前的最新事件。如果这是最近发生的事件，我建议你查阅可靠的新闻源或官方声明以获取最新的信息和准确的分析。 |

[1] 具体事故涉及人员名称可被 ChatSOS 有效抓取，但是受个人隐私保护影响，测试结果中相关数据已被人为省略

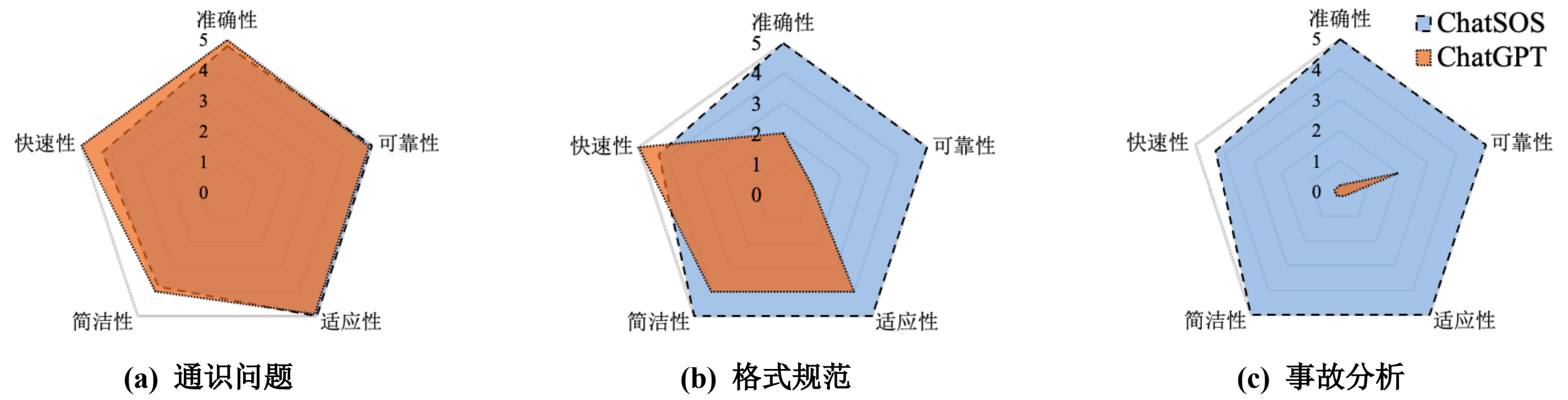

(a) 通识问题　　(b) 格式规范　　(c) 事故分析

图 4　各项评分对比

Fig.4　Comparison of scores across categories

## 4　结 论

1）本文从信息技术快速发展的背景出发，探讨了安全行业内信息的获取和管理问题。在传统关系型数据库面临的数据形式、管理难度等挑战下，提出并实现了基于向量数据库的解决方法，这一方法显著提高了检索的效率和泛化能力，同时基于知识的问答也有效减少了大模型回答中出现幻觉问题的频率。

2）本研究引入了大语言模型，充分发挥其在自然语言处理任务中的强大能力。并针对其在信息时效性和领域特定语料的充足性方面的不足，因此探讨了提示工程技术的有效性。通过为特定需求设计多样化的提示模板，提高了大语言模型在专业领域的应用效果，实现了对模型的优化和定制化应用，

更好地满足了特定领域的需求。

3）构建了 ChatSOS 系统，利用大语言模型作为生成式核心，通过向量数据库与关系型数据库的结合，实现了对安全事故提问的快速响应与深入分析，并在准确性、可靠性和适应性方面表现优秀。未来工作将进一步探索更多实际应用场景，丰富和完善 ChatSOS 系统的工具组件，以验证和拓展本研究的实用性和有效性。

## 参 考 文 献